\documentclass{article}

\usepackage[authoryear,round]{natbib}
\usepackage{arxiv}
\setcitestyle{numbers}

\usepackage[utf8]{inputenc} 
\usepackage[T1]{fontenc}    
\usepackage{hyperref}       
\usepackage{url}            
\usepackage{booktabs}       
\usepackage{amsfonts}       
\usepackage{nicefrac}       
\usepackage{microtype}      
\usepackage{lipsum}
\usepackage{graphicx}
\graphicspath{ {./images/} }
\usepackage{amsmath}      
\usepackage{amssymb}      
\usepackage{natbib}      
\usepackage{graphicx}      
\usepackage{url}       
\usepackage{algorithm2e}      
\usepackage{multirow}
\usepackage{caption} 

\title{Survival Seq2Seq: A Survival Model based on Sequence to Sequence Architecture}

\author{Ebrahim Pourjafari $^{1}$, Navid Ziaei$^{1}$, Mohammad R. Rezaei$^{1}$, Amir Sameizadeh $^{1}$
, Mohammad Shafiee $^{2}$, Mohammad Alavinia $^{3}$, Mansour Abolghasemian $^{1}$ and Nick Sajadi$^{1}$}

\begin{document}
\maketitle
$^{1}$ \quad Ortho Biomed Inc.; nick.sajadi@orthobiomed.ca \\
$^{2}$ \quad Department of Internal Medicine, University Health Network;\\
$^{3}$ \quad KITE, Toronto Rehabilitation Institute, University Health Network;\\

\begin{abstract}
This paper introduces a novel non-parametric deep model for estimating time-to-event (survival analysis) in presence of censored data and competing risks. The model is designed based on the sequence-to-sequence (Seq2Seq) architecture, therefore we name it Survival Seq2Seq. The first recurrent neural network (RNN) layer of the encoder of our model is made up of Gated Recurrent Unit with Decay (GRU-D) cells. These cells have the ability to effectively impute not-missing-at-random values of longitudinal datasets with very high missing rates, such as electronic health records (EHRs). The decoder of Survival Seq2Seq generates a probability distribution function (PDF) for each competing risk without assuming any prior distribution for the risks. Taking advantage of RNN cells, the decoder is able to generate smooth and virtually spike-free PDFs. This is beyond the capability of existing non-parametric deep models for survival analysis. Training results on synthetic and medical datasets prove that Survival Seq2Seq surpasses other existing deep survival models in terms of the accuracy of predictions and the quality of generated PDFs.
\end{abstract}


\section{Introduction}
\label{section:introduction}
The objective of survival analysis is to estimate the hitting time of one or more events in the presence of censored data. In healthcare, such events can include time of death, inception of a disease, time of organ failure, etc. The ability to predict the time or probability of certain events happening to a patient is a valuable asset for medical professionals and clinical decision makers, as it enables them to manage clinical resources more efficiently, make better informed decisions on the treatment they offer to patients, find the best organ donor-recipient matches, etc. 

Electronic health records (EHRs) often have characteristics that pose several challenges to the development of reliable survival models. Some EHRs are longitudinal, i.e., multiple observations per covariate per patient over time are recorded. The survival model must be able to process such measurements and learn from their sequential temporal trends. Medical records, especially longitudinal observations, tend to be highly sparse. Therefore, any reliable survival model must effectively handle missing values, even if the missing rate is extremely high. Dealing with right censored data is another complication of survival analysis models. The right censored data happens often when medical centers lose track of a patient after a certain time, called the censoring time. Survival models must take into account censored data during the training phase. In addition, the presence of competing risk events is another challenge that survival models need to deal with. Having a long-tail distribution is another characteristic of many medical datasets. In such skewed datasets, samples with shorter event times form the bulk of the distribution, while samples with longer event times make only a small portion of the dataset. Following the distribution of such skewed datasets is another challenge for survival and regression models. In fact, many existing survival models cannot accurately predict PDFs with a long time span from medical datasets.

The major challenge of developing survival analysis models is, however, the non-existence of the ground truth for the probability distribution of risk events. Medical records contain the time and type of events that happened to patients. However, the underlying distribution of time-to-event is unknown. This makes developing statistical models or supervised training of machine learning models for survival analysis difficult. Survival models can be divided into two main categories: parametric and non-parametric. Parametric survival models assume a certain stochastic distribution for the dataset, while trying to estimate the parameters of the assumed distribution. On the other hand, non-parametric survival models do not assume any prior distribution for the events. Instead, they try to estimate distributions purely based on the observed relationship between the covariates and the time of events in the dataset. 

To address the aforementioned challenges of developing survival analysis models and to alleviate the shortcomings of the existing survival models, we propose Survival sequence-to-sequence (Seq2Seq). Survival Seq2Seq is a non-parametric multi-event deep model, capable of processing longitudinal measurements with very high missing rates. The accuracy of our model in predicting event times as well as the quality of its generated  probability distribution functions (PDFs) exceeds that of existing survival models. In addition, Survival Seq2Seq performs superbly on skewed datasets. The superiority of our model is backed by the results obtained by training Survival Seq2Seq on synthetic and medical datasets. These results will be provided in the later sections of this paper.     

Our proposed Survival Seq2Seq model has the following key features:
\begin{itemize}
    \item The first layer of the recurrent neural network (RNN)-based encoder network of Survival Seq2Seq is made of Gated Recurrent Units with Decay (GRU-D) \cite{grud} cells. GRU-D cells offer superior performance in imputing not-missing-at-random values. Taking advantage of GRU-D, Survival Seq2Seq can effectively handle high missing rates that commonly occur among medical datasets.
    \item The decoder network of our model is a recurrent network, which can generate substantially smoother PDFs compared to other non-parametric survival models. Since Survival Seq2Seq has fewer trainable parameters compared to a decoder made of dense layers, it suffers less from overfitting compared to other non-parametric models that use Multi-Layer Perceptron (MLP) in their decoders.
    \item We have enhanced the typical loss function used for training non-parametric survival models by improving the ranking loss term of our model. The improved ranking loss will help the model to better rank samples with longer event times. 
    \item Our proposed model can be effectively trained on datasets with a long-tailed distribution, which is a common characteristic of healthcare datasets. This means that Survival Seq2Seq can accurately predict longer event times as well as the shorter ones. 
\end{itemize}

\section*{Generalizable Insights about Deep Survival Models in the Context of Healthcare}
\label{section:healthcareApplications}
The goal of survival analysis is to provide a hazard or reversely a survival function about a medical event for a patient, given some clinical observations. The hazard function or equivalently the hazard rate shows the probability of the occurrence of a medical event such as death or organ failure over time.  While survival analysis has a long history in healthcare and non-healthcare applications using Cox-based proportional hazard (CPH) models, recent publications in the literature show the superior performance of deep learning models for predicting hazard functions. This means that the occurrence rate of medical events can be predicted more accurately over time. Consequently, healthcare providers can take preemptive actions more accurately and adequately.

As an example, the time of death of potential donors can be predicted via hazard function, which allows procurement teams to make a timely attempt. Moreover, a deployed tool based on accurate hazard function can improve matchmaking and the outcome of transplants, leading to shorter wait lists and waiting times, improved longevity of the organs after transplantation, less need for re-transplantation, and longer survival of recipients with a higher quality of life.

Currently, medical decision-making tools based on manual calculations are very complex and not fully supported by solid emerging evidence. For example, it is challenging to accurately predict the death time of the potential organ Donation after Circulatory Death (DCD) donors to allow successful donation. Accordingly, over 30\% of DCD attempts are unsuccessful, and some centers in Canada refuse to implement DCD programs due to their low success rate \cite{iscwebsite}. Therefore, there is an urgent need to implement a precise clinical real-time decision support system for donor evaluation as well as organ suitability. Additionally, the major advantage of machine learning techniques is their ability to adapt to new patient data as it accumulates.

The rest of this paper is organized as follows: A short discussion of related works and their limitation is provided in section \ref{section:relatedwork}. The architecture of Survival Seq2Seq is described in Section \ref{section:seq2seq}. Section \ref{section:cohort} represents the datasets used for training the model, while the experimental results of the training are provided in Section \ref{section:experiments}. Finally, section \ref{section:conclusion} concludes the paper.

\section{ Related Work}
\label{section:relatedwork}
Classical parametric models rely on strong assumptions about the time-to-event distribution. Such strong assumptions allow these models to estimate the underlying stochastic process based on the observed relationship between covariates and time-to-event. However, the predicted probability distribution of these models is over-simplified and often unrealistic. CPH is an example of a parametric statistical model that simplifies the underlying distribution by assuming that the proportional hazard increases constantly over time. The model estimates the hazard function, $\lambda(t|\textbf{x})$, the probability that an individual will experience an event within the time $t$, given all $\textbf{x}$ features (covariates). Although several works such as \cite{vinzamuri2014active,vinzamuri2013cox,li2016multi} have tried to address the shortcomings of the CPH model to some degree, the over-simplification of the underlying stochastic process limits the flexibility, generalizability, and the prediction power of the CPH model. Besides, the CPH model cannot process longitudinal measurements.
Deep Survival Machines (DSM) \cite{dsm2020} is an example of a machine learning-based parametric model that assumes a combination of multiple Weibull and Log-Normal primitive distributions for the survival function. A deep MLP model is trained to estimate the parameters of those distributions and a scaling factor to determine the weight of each distribution in the overall estimated hazard PDF. Training DSM is rather difficult, as the optimizer is easily diverged when trying to minimize its loss function and the model becomes overfit. Despite delivering a high Concordance Index (CI) score, the model performs poorly when estimating the hitting time of events. The CI score is a metric used for evaluating the ranking performance of survival models. The model is also unable to process longitudinal measurements.

The random survival forests (RSFs) method \cite{ishwaran2008random} is an extension of random forests that supports the analysis of right-censored data. The training procedure of RSFs is similar to other random forests. However, the branching rule is modified to account for right-censored data by measuring the survival difference between the samples on either side of the split, i.e., survival times. RSFs have become popular as a non-parametric alternative to CPH due to their less restrictive model assumptions. However, similar to CPH, RSFs cannot be trained on longitudinal datasets.

In \cite{katzman2018deepsurv}, a non-parametric variation of CPH called DeepSurv is proposed as a treatment recommendation system. DeepSurv uses an MLP network for characterizing the effects of a patient’s covariates on their hazard rate. This solves the parametric assumptions of the hazard function in the original CPH model. It also leads to more flexibility of DeepSurv compared to CPH. As a result, DeepSurv outperforms other CPH-based models and it can learn complex relationships between an individual’s covariates and the effect of a treatment. DeepCox \cite{nagpal2021deep} proposes Deep Cox Mixtures (DCMs) for survival analysis, which generalizes the proportional hazards assumption via a mixture model, by assuming that there are latent groups and within each, the proportional hazards assumption holds. DCM allows the hazard ratio in each latent group, as well as the latent group membership, to be flexibly modeled by a deep neural network. However, both DeepSurv and DeepCox models still suffer from the same strong assumption of proportional hazards as the original CPH formulation. Also, neither of those two models supports longitudinal measurements.

As an alternative to assuming a specific form for the underlying stochastic process, \cite{lee2018deephit} proposes a non-parametric deep model called DeepHit, to model the survival functions for the competing risk events. Since no assumptions are made on the survival distribution, the relationship between covariates and the event(s) can now change over time. This is considered an advantage of DeepHit over CPH-based models. The first part of the DeepHit model, i.e., the encoder, is made of a joint MLP block. The decoder of the model is made of MLP blocks, each specific to one event. The output of each case-specific block is a discrete hazard PDF for each event. The last layer of each case-specific block contains $N$ units, where each unit generates the likelihood for one timestep ($e.g.$ one hour or one month) of the hazard PDF over the prediction horizon. A major drawback of this method for generating probability distributions is that the predictions of the output layer could arbitrarily vary from one unit to the next. This causes the overall generated PDFs to be extremely noisy. Moreover, depending on the number of time steps in the prediction horizon, the number of trainable parameters in the output layer could become too high and may cause overfitting when training the model. 

Dynamic DeepHit (DDH) \cite{lee2019dynamic} is an extension of DeepHit, capable of processing longitudinal measurements. In this model, the encoder is replaced with RNN layers followed by an attention mechanism. The RNN block can learn the underlying relationship between longitudinal measurements and provide finer predictions compared to the MLP block in DeepHit. The case-specific blocks of DeepHit and DDH are the same, which means that DDH suffers from predicting noisy PDFs and overfitting.  

Recently, the application of the transformer architecture in survival analysis has been studied in some research such as \cite{wang2021survtrace} and \cite{pmlr-v146-hu21a}. However, none of the models developed in the mentioned research can process longitudinal measurements.

\section{Methods: Survival Seq2Seq}
\label{section:seq2seq}

A successful non-parametric survival model must accept longitudinal EHRs as the input, and must output hazard PDFs for competing risks, while using survival times as the ground truth for training. We realized that the Seq2Seq architecture had a strong potential for performing such a survival analysis. This architecture is commonly used for natural language processing (NLP) tasks. For example, it can translate one language to another. A typical Seq2Seq model is made up of an RNN-based encoder network and another RNN-based network as the decoder. In case of language translation, the encoder encodes a sentence of the source language and sends the encoded sequence to the decoder. The decoder then generates the sentence of the destination language word by word based on the previously generated words in the destination sentence as well as the encoded sequence. Also, adding an attentional mechanism to a Seq2Seq model improves the performance of the model when translating longer sentences. 

The Seq2Seq architecture can be adapted to perform survival analysis. In a nutshell, the encoder of a Seq2Seq model can process longitudinal EHR and send the encoded sequence to the decoder, while the decoder generates a discrete hazard PDF of an event based on the encoded sequence. Despite showing a strong potential, the original Seq2Seq architecture needs to undergo several modifications to be ready for performing survival analysis. The model needs a missing value handling mechanism to effectively impute the sheer number of missing values in longitudinal medical datasets. The decoder of Seq2Seq can generate a PDF for each competing risk. In addition, a proper loss function is required to train such a non-parametric model. While considering censored data, such a loss function must be able to shape the generated PDFs based on the observed relationship between measurements and event times. 

\begin{figure}
  \centering
  \includegraphics[scale=0.20]{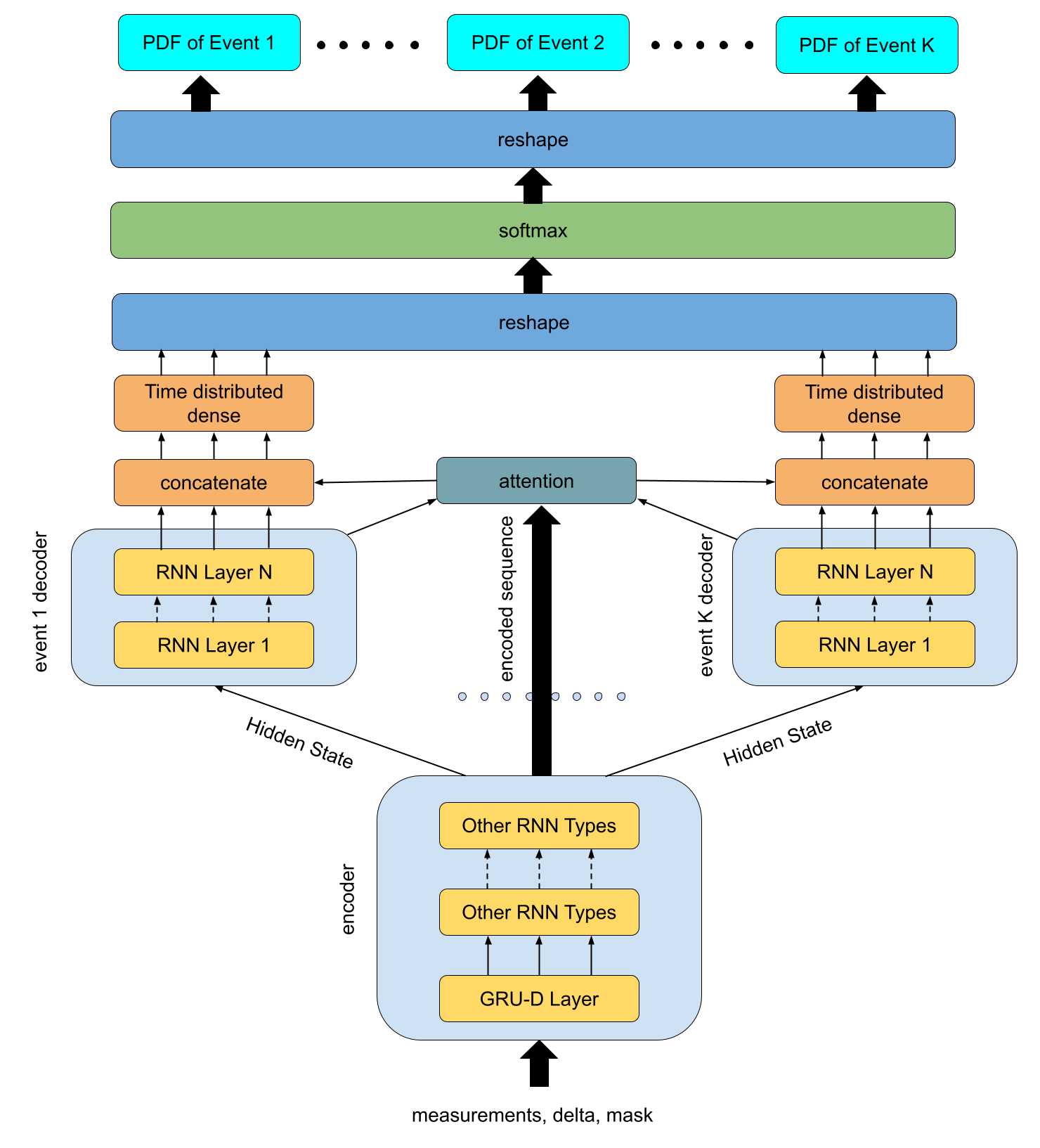}
  \caption{The overall architecture of Survival Seq2Seq}
  \label{fig:seq2seq1}
\end{figure}

Figure \ref{fig:seq2seq1} shows the overall structure of Survival Seq2Seq, a survival model based on the Seq2Seq architecture. The model follows the basic Seq2Seq architecture with an encoder for processing longitudinal measurements and a decoder for generating hazard PDFs for multiple events. The first RNN layer of the encoder is made up of GRU-D cells. These cells are highly effective in imputing missing values of medical datasets. The decoder network is composed of several RNN-based decoder blocks, where each block is responsible for generating the PDF for one event in the dataset. We have added an attentional mechanism to Survival Seq2Seq to improve its overall performance. To train Survival Seq2Seq, we use a multi-term loss function composed of the log-likelihood loss \cite{MLT2006} plus an improved ranking loss term that enhances the ranking performance of the model. Design of Survival Seq2Seq is discussed in greater detail in the rest of this section. 

\subsection{Encoder}
The encoder network is responsible for processing longitudinal measurements of patients and passing the encoded sequence to the decoder network. As mentioned earlier, longitudinal EHRs are very sparse. To handle the missing values of longitudinal measurements, we use GRU-D cells in the first layer of the encoder network. There is often a strong correlation between the missing pattern of covariates and labels in medical datasets. A GRU-D cell learns the relationship between labels and the missing pattern of covariates during a supervised learning process and utilizes the observed relationship to impute missing longitudinal measurements for continuous covariates. GRU-D imputes missing values of a covariate by applying a decay rate to the last measured value of that covariate. The influence of the last covariate measurement is reduced proportionally over time if the covariate has not been measured. If a covariate is not measured for a long period of time, GRU-D imputes that missing value by relying more on the mean value of that covariate rather than its last measurement. The imputation process in GRU-D needs three input vectors: a vector containing longitudinal measurements, a mask vector which indicates if a measurement is available or missing (value one represents a covariate is measured and value zero represents the covariate is missing) and a Delta vector which represents the time difference between the current timestamp and the last timestamp at which a covariate for a patient was measured. Readers can refer to \cite{grud} for the implementation details of GRU-D cells. The encoder network of Survival Seq2Seq can be stacked with multiple RNN layers if desired. In that case, the stacked layers can be made up of some other recurrent cells such as Gated Recurrent Unit (GRU) or Long Short-Term Memory (LSTM). A stacked encoder allows the model to learn more complex temporal relationships in data, provided that the training set is large enough to support a deeper network. If not, the encoder can also work with only one GRU-D layer.

\subsection{Decoder}
As depicted in Figure \ref{fig:seq2seq1}, the decoder network is made of $K$ decoder blocks, each specific to one event in the dataset. The censored event is not considered as an interested event, therefore no decoder block is assigned for censored records. Decoder blocks can be made of stacked RNN layers of vanilla RNN, LSTM or GRU. Similar to the encoder network, a stacked decoder block can be used to learn more complex relationships in data. The generated hazard PDFs are discrete, meaning that the prediction horizon is divided into several timesteps (bins) and a decoder block predicts the value (likelihood) of each bin sequentially. To generate the likelihood for a given timestep, the decoder relies on the likelihood of the previous timestep as well as the encoded sequence. Therefore, the value of the PDF for any timestep is dependent on the previous timesteps and cannot arbitrarily change. As a result, the generated PDFs are smooth and ripple-free.

The output of each decoder block is concatenated with the attention vector and then passed through a time distributed dense layer with a \textit{relu} activation function. The attentional mechanism improves the performance of the model in dealing with long encoded sequences, i.e,  data samples with too many measurements. The output tensor of all decoder blocks is reshaped to a one-dimensional tensor, where a \textit{softmax} activation is applied on the resulted tensor. The \textit{softmax} activation guarantees that the joint cumulative distribution function (CDF) of all events is always equal to one. The output of the \textit{softmax} layer is then reshaped to its original dimensions. This provides the estimated hazard PDFs for all events as follows:
\begin{equation}
\label{eq:seq2seq_output}
PDF = 
 \begin{pmatrix}
  p_{1,1} & p_{1,2} & \cdots & p_{1,T_h} \\
  p_{2,1} & p_{2,2} & \cdots & p_{2,T_h} \\
  \vdots  & \vdots  & \ddots & \vdots  \\
  p_{K,1} & p_{K,2} & \cdots & p_{K,T_h} 
 \end{pmatrix},
\end{equation}
where each row represents the predicted hazard PDF for an event, while $T_h$ is the number of timesteps (length of decoder) for the estimated PDFs. Using Equation \ref{eq:seq2seq_output}, the estimated hazard CDF for event $k^*$ at timestep $\tau$ for a set of covariates $x$ is given by

\begin{equation}
\label{eq:seq2seq_cdf}
CDF_{k^*}(\tau|x) = \sum_{i=0}^{\tau}p_{k^*, i}
\end{equation} 

Depending on the use case, either of the PDF formulated in Equation \ref{eq:seq2seq_output} or the CDF in Equation \ref{eq:seq2seq_cdf} can be considered as the output of the model. Also, if the hitting time of an event is desired, the expected value of the PDF can be considered as the predicted time of the event.

\subsection{Loss function}
The loss function used for training Survival Seq2Seq is $L=L_l+L_r$ in which $L_l$ and $L_r$ are log-likelihood \cite{MLT2006} and ranking terms, respectively. The log-likelihood term is defined as follows:
\begin{equation}
\label{eq:log_likelihood}
    L_l = -\sum_{j \in U_{uc}}log(p_{k_{t},\tau_t}^j) - \sum_{j \in U_c}log(1-\sum_{k=1}^{K}CDF_{k}(\tau_t|x_j)),
\end{equation}
in which $U_{uc}$ and $U_c$ are the sets of uncensored and censored patients, respectively. The index $k_t$ is the ground truth for the first hitting event, while $\tau_t$ is the time of the event or censoring. The log-likelihood loss is the main loss term for training Survival Seq2Seq. The ground-truth for the probability distribution of events is unknown to us. The log-likelihood loss allows a non-parametric model such as Survival Seq2Seq to be trained for predicting probability distributions, while using the first hitting event time as the ground truth.  
The first term of this loss is used for training the model on the first hitting event and its corresponding time for uncensored patients, while the second term is used for training the model on censored data. The first term of $L_l$ is designed to maximize the estimated hazard PDF for event $k_t$ at $\tau_t$, while minimizing it for other timesteps. 

$L_r$ is a ranking loss and improves the ranking capability of the model. The idea behind the ranking loss is that if an event happens for a given patient at a given time, the estimated CDF for that patient at the time of the event must be higher than patients who have not experienced the event yet. Therefore, this loss term increases the overall CI score of the model. According to \cite{Jing2019} the ranking loss can be defined as:

\begin{equation}
\label{eq:ranking1}
L_r = -\frac{1}{|U_a|}\sum_{(i,j)\in U_a}\Phi(CDF_{k}(\tau_{t_i}|x_{i})-CDF_{k}(\tau_{t_i}|x_{j})),
\end{equation}
where $\tau_{t_i}$ is time of event for patient $i$ and $\Phi(.)$ is a differentiable convex function where we use the exponential function $\Phi(z)=exp(z)$.
The $U_a$ term is a set of acceptable patient pairs. In an acceptable pair, the first element must be uncensored, and the event for the first element must occur before the event or censoring time of the second element.
We modify the ranking loss in Equation \ref{eq:ranking1} and propose the following:
\begin{equation}
\label{eq:ranking2}
L_r=-\frac{1}{|U_a|}\sum_{t}\sum_{(i,j)\in U_a}\Phi(CDF_{k}(\tau_t|x_{i})-CDF_{k}(\tau_t|x_{j})).
\end{equation}
The difference between Equation \ref{eq:ranking2} and Equation \ref{eq:ranking1} is that the updated ranking loss compares two elements of a given pair at every timestep over the prediction horizon, while the loss in Equation \ref{eq:ranking1} only compares the pair at the time of the event. Our experimental results provided in the upcoming sections suggest that the updated ranking loss improves the ranking capability of the model, especially for events with longer hitting times. A longer hitting time means more time steps at which the modified ranking loss evaluates an acceptable pair. This translates to a better ranking performance on events with longer hitting times. 

\section{Cohort}\label{section:cohort}
\begin{figure}[!tb]
		\centering
		\includegraphics[scale=0.50]{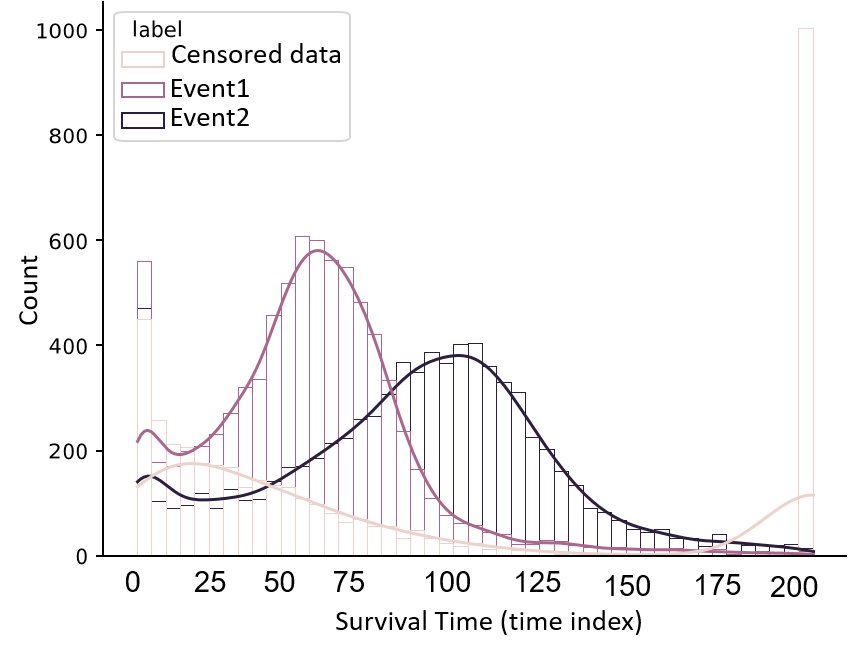}
		\caption{The histogram of survival times in the synthetic dataset. The histogram shows that the generated dataset is long-tailed similar to medical datasets.}
		\label{fig:histogram_events}
	\end{figure}

The datasets used for training the model are described in the following:

\textbf{SYNTHETIC:} 
To evaluate the performance of Survival Seq2Seq, we created a synthetic dataset based on a statistical process.
Here, we consider $\boldsymbol{x}=(x^{0},...,x^{K})$ as a tuple of $K$ random variables in a way that each random variable has an isotropic Weibull distribution
$$ f(x;\gamma,\mu, \alpha) = \frac{\gamma} {\alpha} (\frac{x-\mu} {\alpha})^{(\gamma - 1)}\exp{(-((x-\mu)/\alpha)^{\gamma})} \hspace{.3in} x \ge \mu; \gamma, \alpha > 0, $$
where $\gamma$ is the shape parameter, $\mu$ is the location parameter and $\alpha$ is the scale parameter. We model the distribution of two event times,  $T^{(1)}_i$ and $T^{(2)}_i$, for each data sample $i$ as a nonlinear combination of these $K$ random variables at time index $i$ defined by
\begin{equation}
    T^{(1)}_i = f(\boldsymbol{\alpha}^T\times{(\boldsymbol{x}_{i}^{k_1})}^2+\boldsymbol{\beta}^T\times(\boldsymbol{x}_{i}^{k_2})),
\end{equation}
\begin{equation}
    T^{(2)}_i = f(\boldsymbol{\alpha}^T\times{(\boldsymbol{x}_{i}^{k_2})}^2+\boldsymbol{\beta}^T\times(\boldsymbol{x}_{i}^{k_1})). \ \ \ \ \ k_1,k_2\in \{1,..., K\}
\end{equation}
where $k_1$ and $k_2$ are two randomly-selected subsets of $K$ covariates that satisfy $k_1 \cap k_2 = \varnothing, k_1 \cup k_2 = K$. By selecting the Weibull distribution in our synthetic data generator, the event times will be exponentially distributed with an average that depends on a linear (with parameters set $\boldsymbol{\beta}$) and quadratic (with parameters set $\boldsymbol{\alpha}$) combination of the random variables. This means long-tailed distributions for the event times similar to medical datasets, as shown in Figure \ref{fig:histogram_events}. The figure shows that the number of samples in the histogram are not monotonically decreasing over the time. Instead, the number of samples for each event peaks at some time and then decreases with a long tail. We chose this deliberately to evaluate the performance of the model on a dataset with a complex time-to-event distribution. For each data sample, we have $(\boldsymbol{x}_i, \delta_i, T_i)$, where $\delta_i=min\{T^{(1)}_i,T^{(2)}_i \}$ identifies the type of the event happening at $T_i$, i.e.,  the event time.
In sum, we considered $K=20$ and generated 20000 data samples from the defined stochastic process with 20\% censoring rate, and a not-missing-at-random rate of 77\% and a maximum event time of 200. To simulate longitudinal measurements, we assume that the covariates for each sample are measured a random number of times, while the measured values at each time stamp are increased or decreased nonlinearly with a random rate specific to each sample-covariate pair. 

\textbf{MIMIC-IV:} The MIMIC-IV dataset \cite{johnson2020mimic} contains hospital clinical records for patients admitted to a tertiary academic medical centre in Boston, MA, USA., between 2008-2019. This database contains demographics data, laboratory measurements, medications administered, vital signs and diagnosis of more than 71000 patients with an in-hospital mortality rate of about $12\%$. We considered a 33-day prediction horizon with a 4-hour time resolution for the decoder of Survival Seq2Seq. Therefore, patients with an event time longer than 33 days were considered censored. We also selected 108 covariates from this dataset based on the feedback of our medical team as well as by applying feature selection methods to the dataset. 

\section{Results}\label{section:experiments}
The performance of Survival Seq2Seq was evaluated by training the model on SYNTHETIC and MIMIC-IV datasets. Results were compared to DDH for benchmarking, as DDH was the only survival model known to us, capable of processing longitudinal measurements for survival analysis. 
\subsection{Evaluation Approach/Study Design}

\raggedright 
Mean absolute error (MAE) is the main metric we use for evaluating the performance of the models. MAE is defined as the mean of the absolute difference between the predicted time of an event and the observed time for that event for uncensored data samples. We consider the expected value of a predicted PDF as the predicted event time. The other metric used in this paper is the time-dependent CI score \cite{antolini2005time} defined as follows:
$$
\mathbb{CI}(t)=P(\hat{F}(t|x_i) > \hat{F}(t|x_j)| \delta_i =1, T_i<T_j, T_i\leq t),
$$
where $\hat{F}(t|x_i)$ is the estimated CDF of the event, truncated at time $t$, given a set of covariates $x_i$. This metric evaluates the performance of the models when predicting the order of events for data samples. The CI measure can evaluate the performance on both censored and uncensored samples in the test dataset. The time dependency of this metric allows us to evaluate the performance of models when capturing the possible changes in risk over time. 

The results will be reported in four 25\%, 50\%, 75\% and 100\% quantiles with 5-fold cross-validation. We consider the length of the decoder of Survival Seq2Seq 25\% longer than the maximum event time of each dataset. This is necessary for dealing with censored samples at the maximum event time (200 for SYNTHETIC and 33 days for MIMIC-IV) or a time close to the defined maximum event time. This extra decoder length allows the model to shift a considerable portion of the PDF for those samples to later timesteps, so that the predicted time of the event for censored data happens after the censoring time.

\subsection{Evaluating Using the SYNTHETIC Dataset}
\begin{figure}
  \includegraphics[width=\linewidth]{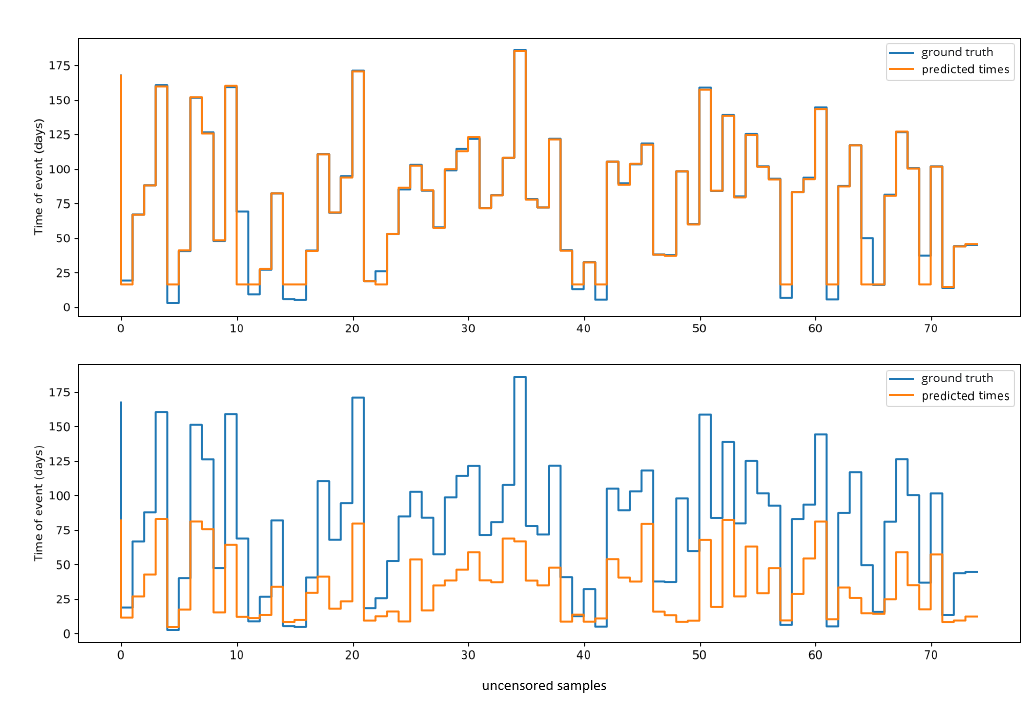}
  \caption{The difference between the ground truth and predictions of Survival Seq2Seq (top) and DDH (bottom) on a few uncensored samples from the SYNTHETIC dataset. The comparison shows that Survival Seq2Seq follows the distribution of the event times more accurately than DDH. The difference between the performance of the two models is more apparent for longer event times.}
  \label{fig:distribution}
\end{figure}

Table \ref{tab:synthetic_mae} compares the performance of Survival Seq2Seq and DDH on the SYNTHETIC dataset in terms of MAE. The mean and variance of the results of the five folds are used for calculating the confidence interval of predictions. It can be seen that Survival Seq2Seq significantly outperforms DDH for both events in all quantiles except the first quantile. Figure \ref{fig:distribution} represents the difference between the ground truth and prediction for Survival Seq2Seq and DDH for a few uncensored samples of the SYNTHETIC dataset. It can be observed that Survival Seq2Seq can follow the distribution of the dataset by predicting values close to the ground truth, whether the survival time is closer to zero or it is much longer. On the other hand, DDH performs very poorly in predicting longer event times. Prediction results provided in Table \ref{tab:synthetic_mae} as well as Figure \ref{fig:distribution} proves that Survival Seq2Seq can effectively predict the event time on datasets with long-tailed distributions.

As Figure \ref{fig:distribution} reveals, DDH has a tendency to predict shorter event times for all events, whether the ground truth is long or short. This translates to a lower MAE for shorter event times for DDH compared to Survival Seq2Seq. This situation is somehow similar to an imbalanced binary classification in which a naively-trained classification model leans toward predicting in favor of the class with the higher share of training samples. The overall accuracy of such a classification model could be very high, although the model would perform very poorly on the minority class. Using the same analogy, a better way to evaluate a model like DDH with skewed predictions is to look at its performance in the last quantile, where events with longer hitting times can be found. This is where Survival Seq2Seq outperforms DDH.  

The slightly higher MAE of Survival Seq2Seq compared to DDH in the first quantile can also be explained based on the imputation mechanism of GRU-D cells. The number of measurements for data samples with very short event times are scarce. If the value of a covariate for a given data sample is missing at the first measurement timestamp, GRU-D has to rely solely on the mean of that covariate to impute that missing value. This is obviously not ideal and causes higher prediction errors for very early predictions. However, the imputation performance of GRU-D improves as more measurements are accumulated.

\begin{table}
\centering
\begin{tabular}{cc|c|c|c|c|c}
\cline{3-6}

& & \multicolumn{4}{ c| }
{Quantiles} 
\\ \cline{3-6}
& & 25\% & 50\% & 75\% & 100\% \\ \cline{1-6}
\multicolumn{1}{ |c  }
{\multirow{2}{*}{Survival Seq2Seq} } &
\multicolumn{1}{ |c| }
{event 1} & 11.85$\pm 0.6$ & 12.47 $\pm$ 1.2 & 14.01 $\pm$ 1.4 & 15.54 $\pm$ 1.8 & 
\\ \cline{2-6}
\multicolumn{1}{ |c  }{}   &
\multicolumn{1}{ |c| }{event 2} & 12.26 $\pm$ 0.5 & 13.20 $\pm$ 0.9 & 15.55 $\pm$ 2.2 & 20.83 $\pm$ 3.8 & \\
\cline{1-6}
\multicolumn{1}{ |c  }{\multirow{2}{*}{DDH} } &
\multicolumn{1}{ |c| }{event 1} & 8.79 $\pm$ 0.7 & 20.36 $\pm$ 1.8 & 29.93 $\pm$ 1.8 & 35.03 $\pm$ 1.8 & \\ \cline{2-6}
\multicolumn{1}{ |c  }{}                        &
\multicolumn{1}{ |c| }{event 2} & 10.32 $\pm$ 0.3 & 18.45 $\pm$ 1.4 & 33.96 $\pm$ 1.5 & 52.22 $\pm$ 1.2 & \\ 
\cline{1-6}

\end{tabular}

\caption{Comparison between the MAE of Survival Seq2Seq and DDH on the SYNTHETIC dataset. Results are reported with 95\% confidence interval.}
\label{tab:synthetic_mae}
\end{table}

\raggedright 
Table \ref{tab:synthetic_ci} represents the time-dependent CI score for the two models on the SYNTHETIC dataset. Despite providing marginally lower CI scores for the two events in the first quantile, our model outperforms DDH in all other quantiles, with a significant 0.08 higher average CI score compared to DDH on the last quantile. The higher CI score of Survival Seq2Seq with respect to DDH can be contributed to the introduction of the new ranking loss in Equation \ref{eq:ranking2}, which improves the ranking performance of the model.

\begin{table}
\centering
\begin{tabular}{cc|c|c|c|c|c}
\cline{3-6}
& & \multicolumn{4}{ c| }{Quantiles} \\ \cline{3-6}
& & 25\% & 50\% & 75\% & 100\% \\ \cline{1-6}
\multicolumn{1}{ |c  }{\multirow{2}{*}{Survival Seq2Seq} } &
\multicolumn{1}{ |c| }{event 1} & 0.835$\pm$0.04 & 0.774$\pm$0.06 & 0.827$\pm$0.02 & 0.844$\pm$0.02 &     \\ \cline{2-6}
\multicolumn{1}{ |c  }{}                        &
\multicolumn{1}{ |c| }{event 2} & 0.850$\pm$0.03 & 0.804$\pm$0.04 & 0.808$\pm$0.03 & 0.835$\pm$0.03 &     \\
\cline{1-6}
\multicolumn{1}{ |c  }{\multirow{2}{*}{DDH} } &
\multicolumn{1}{ |c| }{event 1} & 0.842$\pm$0.01 & 0.766$\pm$0.01 & 0.760$\pm$0.01 & 0.776$\pm$0.01 & \\ \cline{2-6}
\multicolumn{1}{ |c  }{}                        &
\multicolumn{1}{ |c| }{event 2} & 0.853$\pm$0.01 & 0.787$\pm$0.01 & 0.744$\pm$0.01 & 0.743$\pm$0.01 & \\ 
\cline{1-6}
\end{tabular}
\caption{Comparison between the time-dependent CI score of Survival Seq2Seq and DDH on the SYNTHETIC dataset. Results are reported with 95\% confidence interval.}
\label{tab:synthetic_ci}
\end{table}

\subsection{Evaluating Using the MIMIC-IV Dataset}
\raggedright 
The performance of Survival Seq2Seq on MIMIC-IV is provided in Table \ref{tab:mimic_scores}. This table shows the MAE and CI scores for our model and compares them to the outcome of DDH. One can observe a pattern similar to Table \ref{tab:synthetic_mae} when comparing the MAE of Survival Seq2Seq and its counterpart. The MAE of our model is marginally higher than DDH in the first quantile. However, Survival Seq2Seq beats DDH in all other quantiles, so that the mean absolute error of our model is less than half of the MAE of DDH on 100\% of the test data. Besides, the performance of Survival Seq2Seq in terms of the CI score exceeds the outcome of DDH for all quantiles. Again, the updated ranking loss in Equation \ref{eq:ranking2} is a factor that contributes to these high CI scores.

\begin{figure}
  \includegraphics[scale=0.50]{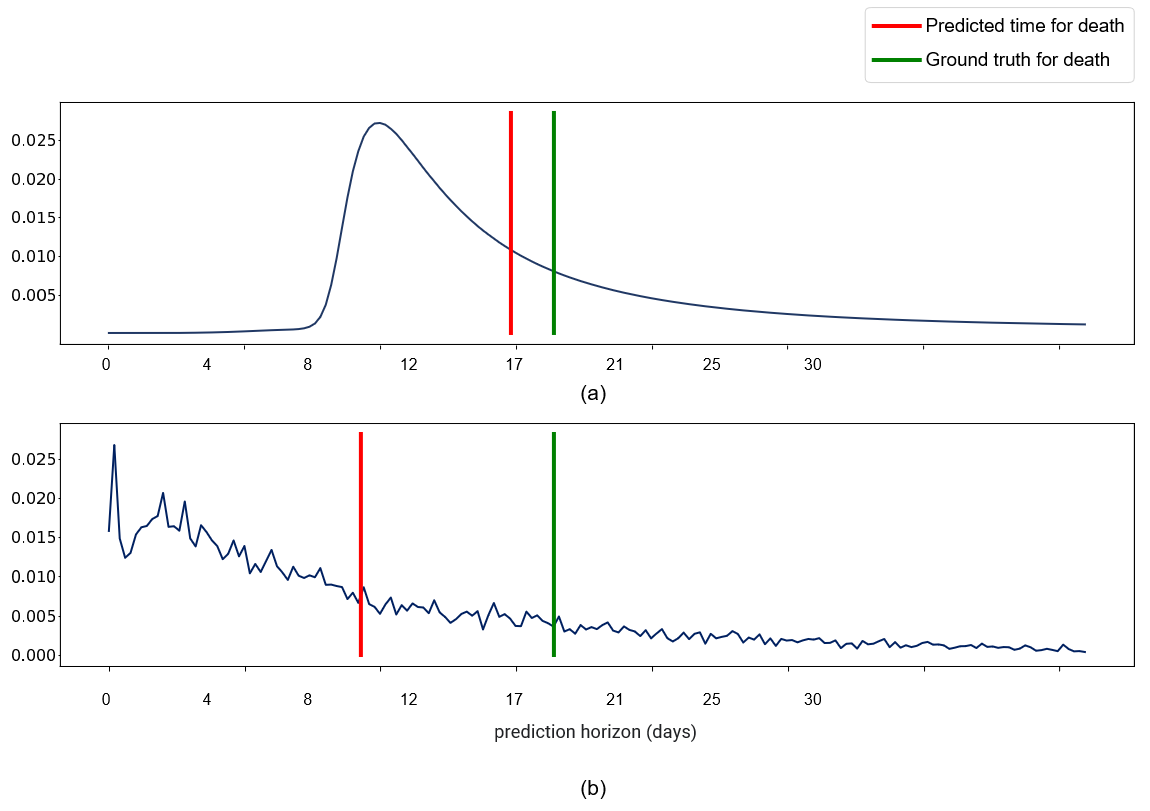}
  \caption{(a) The PDF generated using Survival Seq2Seq for a random uncensored patient of MIMIC-IV and (b) The generated PDF of DDH for the same patient. The expected value of each PDF is considered as the predicted time of death. The comparison shows that the quality of the predicted probability distribution of Survival Seq2Seq is superior than DDH.}
  \label{fig:PDFs}
\end{figure}

The quality of generated PDFs is another important factor besides the prediction accuracy when comparing Survival Seq2Seq to other non-parametric models. The superiority of our model becomes apparent from Figure \ref{fig:PDFs}, where the predicted PDFs of Survival Seq2Seq and DDH are compared for a random uncensored patient in MIMIC-IV. As described in earlier sections, the RNN-based decoder of our model can generate smooth and ripple-free probability distributions. This is in contrast to DDH where using an MLP-based decoder results in PDFs with high fluctuations, as shown in Figure \ref{fig:PDFs}.

\begin{table}
\centering
\begin{tabular}{cc|c|c|c|c|c}
\cline{3-6}
& & \multicolumn{4}{ c| }{Quantiles} \\ \cline{3-6}
& & 25\% & 50\% & 75\% & 100\% \\ \cline{1-6}
\multicolumn{1}{ |c  }{\multirow{2}{*}{Survival Seq2Seq} } &
\multicolumn{1}{ |c| }{MAE} & 34.83$\pm$4.1 & 37.06$\pm$4.6 & 39.53$\pm$4.0 & 62.74$\pm$3.2 &     \\ \cline{2-6}
\multicolumn{1}{ |c  }{}                        &
\multicolumn{1}{ |c| }{CI} & 0.876$\pm$0.02 & 0.882$\pm$0.02 & 0.885$\pm$0.02 & 0.906$\pm$0.02 &     \\
\cline{1-6}
\multicolumn{1}{ |c  }{\multirow{2}{*}{DDH} } &
\multicolumn{1}{ |c| }{MAE} & 32.88$\pm$8.4 & 40.23$\pm$2.4 & 61.17$\pm$2.8 & 125.95$\pm$4.7 & \\ \cline{2-6}
\multicolumn{1}{ |c  }{}                        &
\multicolumn{1}{ |c| }{CI} & 0.863$\pm$0.03 & 0.849$\pm$0.02 & 0.845$\pm$0.03 & 0.528$\pm$0.04 &     \\
\cline{1-6}
\end{tabular}

\caption{Comparison between the results of Survival Seq2Seq and DDH on MIMIC-IV. Results are reported with 95\% confidence interval.}
\label{tab:mimic_scores}
\end{table}

\section{Discussion}
\raggedright 

\label{section:conclusion}
Each of the MAE and CI metrics used in this paper is useful for evaluating a different use case of a survival model. MAE is useful for evaluating the outcome of a survival model when predicting the time of events (calibration) is desired, while CI is a better choice when the model is applied on ranking problems. However, we cannot train a model that provides optimal MAE and CI at the same time. The log-likelihood loss has a higher influence on the calibration capability of the model than the ranking loss. Therefore, to minimize MAE, one must assign a higher weight to the log-likelihood loss than the ranking loss. On the other hand, the ranking loss must be assigned a higher weight if the ranking capability of the model and a higher CI score is desired. Since we believe that calibration is a more important task than ranking, we assigned a higher weight to the log-likelihood loss to minimize MAE. We did the same for DDH to make the comparison between the two models fair. Assigning a higher weight to the ranking loss results in a maximized CI score. For example, we achieved a CI score of 0.932$\pm$0.01 on the last quantile of MIMIC-IV with Survival Seq2Seq when a higher weight was assigned to the ranking loss. This is almost 3\% higher than the CI score reported in Table \ref{tab:mimic_scores} when calibration was the objective.

Survival Seq2Seq can accurately follow the distribution of the event time for long-tail datasets, while DDH and many other survival or regression models cannot. This is an interesting feature of our model and we mainly consider this as a side contribution of the RNN-based decoder network to our model. We were initially not sure if this feature of the model is a result of using GRU-D cells in the encoder of our model, the modified ranking loss, or the use of an RNN-based decoder. We created a second version of Survival Seq2Seq with the same RNN-based decoder network as the original model, but without the GRU-D layer in the encoder and without the modified ranking loss. We used a simple masking technique for dealing with missing measurements. This version of Survival Seq2Seq was still able to follow the distribution of the time-to-event on long-tail datasets, although the accuracy of the model dropped. Therefore, we concluded that the RNN-based decoder was the main contributor to this interesting feature of our model in performing well on long-tail datasets. However, the exact underlying mechanism of this phenomenon is unknown to us and requires further investigation.

Based on the provided results in terms of MAE and CI score, we proved that Survival Seq2Seq has a high prediction accuracy. However, accuracy metrics are not the only factor that must be taken into account when evaluating the performance of a survival model. We believe that the quality of predicted PDFs is as important as the accuracy metrics. As the provided PDF sample in this paper shows, the generated PDFs of Survival Seq2Seq are exceptionally smooth and ripple-free. This is an outstanding feature for a non-parametric survival model. 

\subsection*{Limitations}
The GRU-D layer that makes the first layer of the encoder of our model relies on previously measured values and the mean of covariates to impute missing data. If a covariate for a row of data (a patient) is missing while it is not previously measured, GRU-D has no choice but to impute that missing value with the mean of the covariate. This is not an ideal behavior and we believe that it is contributing to lower performance of Survival Seq2Seq in the first quantile of data, where there are not enough measurements for GRU-D to properly impute missing variables. We acknowledge this limitation of our model on early predictions and encourage the readers to think of ways that could mitigate this phenomenon.

There is a limit to the length of the encoder and decoder of Survival Seq2Seq. Both  networks are RNN-based. Consequently, they cannot pass information through time if their length exceeds a certain limit. However, using the attention mechanism solves this problem to a certain extent. We kept the maximum length of the encoder up to 60 time steps for MIMIC-IV. During data pre-processing, if two covariates of a data sample are measured at relatively close timestamps, we consider the average of those timestamps as the unique timestamp for those two measurements. This helps us to reduce the number of longitudinal measurements for a given row of data (or a patient) in the dataset. If the number of measurements exceeds the maximum of time steps, we only keep 60 randomly-selected measurements from that row of data, while discarding the rest. Although, we were able to use a higher maximum length, we did not notice a meaningful accuracy improvement using an encoder longer than 60 steps on MIMIC-IV. Similarly, the length of the decoder is limited as well. However, we do not believe that the limit on the length of the decoder causes an issue for practical use cases. For example, in our experiments, we successfully trained Survival Seq2Seq on MIMIC-IV with a 2-hour time resolution. The high resolution decoder had a length twice the length of the decoder used for reporting the results in this paper. Such an experiment shows that the length of the RNN-based networks does not limit the ability of Survival Seq2Seq for practical use cases. However, one must be careful not to cause vanishing gradients by setting very lengthy encoder and decoder networks.
\newpage

\bibliographystyle{plainnat}


\end{document}